\documentclass[10pt,journal,compsoc]{IEEEtran}

\ifCLASSOPTIONcompsoc
  \usepackage[nocompress]{cite}
\else
  \usepackage{cite}
\fi

\usepackage[noend]{algpseudocode}
\usepackage[ruled,vlined,linesnumbered]{algorithm2e}
\algrenewcommand{\Return}{\State \textbf{return}\ }
\algnewcommand{\funccall}[1]{\textit{#1}}
\ifCLASSINFOpdf
\else
\fi
     \usepackage{graphicx}  
     \usepackage{float}
\usepackage{graphicx}
\usepackage{caption}
     \usepackage{subfigure}     
             \usepackage[table, dvipsnames]{xcolor}
             
             \usepackage{multicol}

\begin{document}
%
\title{Enterprise Analytics using Graph Database and Graph-based Deep Learning}
%
%
%
%

\author{Shagufta Henna,~\IEEEmembership{Senior Member~IEEE,}
               Shyam Krishnan Kalliadan
\IEEEcompsocitemizethanks{\IEEEcompsocthanksitem Shagufta Henna and Shyam Krishnan Kalliadan are with the Department
of Computing, Letterkenny Institute of Technology, Co. Donegal.\protect\\
E-mail: shaguftehnna@gmail.com

}
\thanks{Manuscript received April 19, 2005; revised August 26, 2015.}}

%
%

\markboth{Journal of \LaTeX\ Class Files,~Vol.~14, No.~8, August~2015}%
{Shell \MakeLowercase{\textit{et al.}}: Bare Advanced Demo of IEEEtran.cls for IEEE Computer Society Journals}
%



\IEEEtitleabstractindextext{%
\begin{abstract}
In a business-to-business (B2B) customer relationship management (CRM) use case, each client is a potential business organization/company with a solid business strategy and focused and rational decisions. This paper introduces a graph-based analytics approach to improve CRM within a B2B environment. In our approach, in the first instance, we have designed a graph database using the Neo4j platform. Secondly, the graph database has been investigated by using data mining and exploratory analysis coupled with cypher graph query language.  Specifically, we have applied the graph convolution network (GCN)  to enable CRM analytics to forecast sales. This is the first step towards a GCN-based binary classification based on graph databases in the domain of  B2B CRM. We evaluate the performance of the proposed GCN model on graph databases and compare it with Random Forest (RF), Convolutional Neural Network (CNN), and Artificial Neural Network (ANN). The proposed GCN approach is further augmented with the shortest path and eigenvector centrality attribute to significantly improve the accuracy of sales prediction. Experimental results reveal that the proposed graph-based deep learning approach outperforms the Random Forests (RsF) and two deep learning models, i.e., CNN and ANN under different combinations of graph features.
\end{abstract}

\begin{IEEEkeywords}
Graph-based deep learning, Business analytics, Graph database for CRM, Enterprise analytics 
\end{IEEEkeywords}}

\maketitle

\IEEEdisplaynontitleabstractindextext

%
\IEEEpeerreviewmaketitle

\ifCLASSOPTIONcompsoc
\IEEEraisesectionheading{\section{Introduction}\label{sec:introduction}}
\else
\section{Introduction}
\label{sec:introduction}
\fi

\IEEEPARstart{G}raphs are visual representations to manage and store complex data beyond the 360-degree overview of business instances. In the context of B2B, it is possible to capture meaningful insights and relationships in CRM with the help of existing data management frameworks and analytic approaches. A CRM system can significantly improve customer experience and brand awareness by using graph-based analytics. The popularity of CRM data lakes has opened several opportunities to improve business performance by using graph storage representation and graph analytics. One of the major dimensions worth to be investigated in B2B is innovative CRM to enhance customer experiences. However,  a cost-efficient solution for improved customer experiences to a business value chain is a challenging task.  It is fascinating to model the CRM system using data-driven strategies to address customer-centric goals. These solutions should extract, store, analyze, and predict potential customers and business opportunities to deliver business value.

Recently, there are some trends to use machine learning for CRM  analytics with improved efficiency. In  \cite{ZhanB2019}, semi-supervised spectral clustering is proposed to analyze customers using sentimental analysis. In \cite{Yang2018}, a recommender system is designed and developed for B2B sales and marketing purposes.  Another recommender system based on the graph database and graph analysis is presented in \cite{TKonno2017}. According to Berkshire Hathaway’s business wire market review, CRM market share forecast is  14.9 \% CAGR between 2020 and 2025 \cite{CRM2020}. Therefore extracting, storing, and analyzing B2B sales and customer data requires more attention. Despite the significant benefits of a graph database and graph analytics in other domains, it has not been investigated in the B2B CRM system.

CRM has evolved into social CRM (SCRM), i.e., web-based social networking for CRM. An SCRM consists of data extraction and data analysis modules. The data extraction module loads relevant data from various sources with the assistance of a crawler. The data analysis module performs knowledge analysis, authority analysis, sentiment analysis, and prediction \cite{Ibrahim2017}. Organizations are increasingly adopting CRM solutions with integrated business intelligence using artificial intelligence and big data solutions to address their strategic needs. Natural language processing \cite{Nugmanova2019}, sentimental analysis \cite{kim2019}, and customer behavior analysis \cite{Asniar2019} are the key performance requirements of a successful CRM analytical module. In \cite{Brahim2020}, the authors highlight the importance of machine learning in a hotel CRM. The proposed work demonstrates the benefits of machine learning and data mining to enhance customer loyalty and hospitality experience. 


In another work in \cite{Taleb2020}, the authors have investigated the benefits and challenges associated with the big data CRM platforms to improve customer experiences and business strategy forecasting.  According to \cite{Kitchin2014}, authors have investigated various risk factors associated with the quality of unverified data for marketing analysis. Another work in \cite{Taleb2020} discusses the vital role of big data platforms to identify business opportunities and business strategy planning.  It also highlights the importance of online feedback, and analysis of competitors' performance, and strategy for product and service offerings in a  successful CRM module.

A relational database management system (RDBMS) cannot capture meaningful and critical relationships essential to devise an intelligent enterprise CRM solution. A graph database can play a vital role here to capture meaningful relationships. Graph databases are NoSQL databases that use various approaches for storage and retrieval of data other than tabular schema-based relations.  Numerous popular enterprise graph database platforms are available in the market, such as ArangoDB, Neo4j, Amazon Neptune, and Orient DB \cite{Das2020}.  Neo4j is an open-source, secure, versatile, and simple graph database platform suitable for enterprise level. It exploits the potentials of built-in cypher graph query language for data analysis and graph data mining. The Neo4j  graph models are scalable, robust, and are Atomicity, Consistency, Isolation, Durability (ACID)-compliant \cite{Huang2013}. Neo4j has a flexible schema to support efficient graph querying, graph mining, and graph visualization through built-in HTTP-API.

In a work in \cite{Kimura2011}, authors used a graph database to model customers to analyze the client requirements with the centrality algorithm. The proposed graph database model identifies the unique customer needs using common links. A business graph is introduced in \cite{McClanahan2016} to model different customer relationships in B2B. The work exploits the power of the graph database to capture these relationship strengths as edge weights between companies represented as nodes. The work uses centrality and spectral analysis to extract the geographical positioning pattern that influences the mutual customer strength.  This work, however, does not consider other important graph factors such as eigenvector centrality, and page rank values.

Recent studies demonstrate that graph analytics based on deep learning has the potential to extract meaning patterns and insights from graph models, e.g., GCN \cite{Gheisari2017}. Deep learning approaches are efficient to extract temporal, spectral, and spatial characteristics and relationships from the graph data. Scarselli et al. \cite{Scarselli2009} introduced the concept of graph neural networks (GNN) \cite{Scarselli2009} based on the neural network concept on directed graphs. GNNs are capable to model complex relationships or dependencies among the nodes of a graph. The spatial version of a GCN was originally implemented in \cite{Micheli2009} where authors applied the message passing principle of GNN on sequential graph layers to reinforce network collective dependency in the spatial domain. In recent work in \cite{Masci2018}, authors have proposed a spatial graph convolution model for meshed surfaces. On contrary to the spatial GCN, the spectral GCNs are based on Fourier transform and aggregates the neighborhood node information \cite{Bruna2014}.  In a work in \cite{Henaff2015}, authors have proposed a GCN model using a  smoothening kernel and parametric filters. This results in a number of learnable features independent of the size of the graph. Another recent work in  \cite{Defferrard2017}, enables explicit learning of laplacian eigenvectors in GCN using polynomial convolution operation. The above studies suggest that there few works in the domain of B2B to use the graph database to model relationships at the enterprise level. However, there are no studies to exploit the potential of graph convolutional networks-based learning on the top of the graph database to model CRM in a B2B.

We propose a novel approach to create a graph database model based on B2B sales CRM dataset to capture all the meaningful dataset features. To achieve this objective, various graph models have been designed and tested to support a variety of data mining and querying capabilities. The work brings forward an efficient and robust graph database management and processing framework on top of Neo4j platform integrated with the B2B sales dataset. In addition, we adopt a graph-based deep learning approach, i.e., GCN  as the first effort on B2B sales graph database model for sales prediction. The GCN model's result provides evidence of how graph-based deep learning outperforms RFs and deep learning solutions such as CNN and feed-forward neural networks (FNNs).
The contributions of this paper are  summarized as follows:

\begin{itemize}
\item  We design a B2B graph database model using a sales dataset based on the Neo4j platform.  The graph database model provides valuable insights and knowledge using graph data mining and graph analytics.

\item We implement graph-based analytics using GCN which illustrates the effectiveness of learned graph features for sales prediction to perform classification.

\item  We show that our proposed B2B CRM graph database coupled with the GCN outperforms RFs and other deep learning approaches such as CNN and FNN under different combinations of graph features extracted from the graph database model.
\end{itemize}

The remaining of this paper is organized as follows.
Section \ref{rw} critically analyzes the related work. The details of the B2B dataset used for this work are presented in \ref{dataset}. Section \ref{crmmodel} presents the design of the graph database model. Feature engineering required for the GCN-based deep learning is discussed in Section\ref{fe}. Section \ref{gcnn} provides a detailed discussion of the proposed GCN-based analytics. Section \ref{pe} evaluates the proposed GCN-based model based on the graph database. It presents the inventory and exploratory data analysis using cypher query. It also details the results and analysis of the GCN model with various graph features. Finally, Section \ref{conc} concludes our work with possible future directions.

\section{Related Work} \label{rw}
This section investigates different related works to our work, i.e., graph databases and graph-based deep learning. 

\subsection{Graph Databases}
Traditional databases and data warehouses cannot process or store unstructured data. NoSQL databases and Hadoop distributed storage systems are alternative solutions in such a scenario. NoSQL database framework, however, requires efficient graph data processing where a graph represents real-life entities and relationships between the entities in a more meaningful manner. Graph representation of big data has significant applications in genetics, social networks, molecular chemistry, finance, and drug testing \cite{Huang2013}. A graph database based on a graph model can store, process, and perform efficient graph analysis such as graph-based deep learning.

A work in \cite{Huang2013} analyzes the performance of Neo4j cypher query performance in various B2B use cases. Neo4j stores the graph data as a record file. It uses a two-layer catching mechanism for faster data retrieval and visualization. The first mechanism stores the reference relationship of nodes with minimal information in the file system cache. The second mechanism stores major graph connectivity and node attribute information in the object cache. Authors in \cite{Das2020} analyze the performance of different NoSQL databases with a suggestion of Neo4j and ArangoDB as the best models. Authors analyze the performance of different NoSQL databases with a suggestion of Neo4j and ArangoDB as the best models. Neo4j-based graph analysis and data mining have applications in various domains. In \cite{Hoksza2015}, Neo4j is used for data mining on various protein graphs. Despite its significant benefits, Neo4j has performance limitations in terms of graph model complexity concerning dataset size. Similarly, the performance of cypher query is limited for the larger subgraphs with a higher degree of connectivity.

Neo4j graph database analysis has also been investigated in the healthcare domain  (Zhao et al., 2019).  In a work, disease prediction is proposed based on the graph model in Neo4j by using Cypher query. Cypher queries can extract hidden patterns to reveal meaningful insights into a business trend. A work in \cite{Zouy2020} analyzes the air crash incidents from 1908 onwards data using Neo4j and Cypher data mining. This model compares the performance of various data import methods in Neo4j. Among the various methods used for import operation, batch-wise import and Neo4j admin import are suggested as the fastest method for large datasets. Another work in \cite{Needham2019} extracts and analyzes various hidden patterns and relationships in the file dataset using Neo4j graph database platform and Cypher query. 

The concept of graph feature representation and analysis for CRM based on social network visual analytic tool VisCRM is introduced in \cite{ye2008}. The proposed model extracts hidden features among the customers in a social network with the help of a graph database model. In its practical application, the model is limited to visual graph analysis and exploratory analysis and lacks predictability capacity. Another application of Neo4j is CRM for goods recommendations using retail knowledge graph and jess reasoning engine \cite{TKonno2017}. The work constructs a graph model based on a retail ontology that is queried and analyzed using Neo4j framework with recommendations offered by the jess reasoning engine. According to \cite{Wangl2014}, a social network-based enterprise relationship graph can deliver higher customer value and business success rates. 

A work in \cite{Aasman2017} investigates knowledge graph based on customer data and CRM with a 360-degree view of a business's client data catalog. The work discusses the concept of customer knowledge graph based on operational knowledge graph, product knowledge graph, and service knowledge graph with an analysis of enterprise data lake. In \cite{Saha2018}, the authors implement a spectral-based graph clustering on the telecommunication call records dataset. The authors compare spectral graph clustering against the K-mean algorithm based on selected features.  Authors in \cite{ZhanB2019}  propose a spectral clustering method based on sentiment lexicon using the social network graph model.  In their work, the authors implement three specific models for constructing a topic-specific sentiment lexicon; a filtering text model, a sentiment relationship graph model, and a spectral clustering model. The proposed model using spectral clustering outperforms the traditional lexicon models whose sentiment analysis performance decreases with an exponential increase in social network text size.

\subsection{Graph-based Deep Learning}
Deep learning based on graph database can be divide into two types: graph representation learning and graph analysis. Deep learning can perform node level and graph-level analysis techniques to identify latent features and knowledge from the data \cite{Zhang2020}.  Graph level operations include graph classification, matching, and graph generation. Examples of node-level operations include node clustering, node recommendation, link prediction, node prediction, and retrieval. Graph representation learning \cite{Cai2020}  is the process of structural data encoding of a graph network. The encoded information is mapped to a low dimensional vector space like the adjacency matrix or Laplacian matrix. These matrices can be directly used in machine learning and data analysis operations. Graph representation learning \cite{Cai2020} is the process of structural data encoding of a graph network. An example of graph representation learning using GNN and its variations is presented in \cite{Scarselli2009}.

Different versions of graph convolutional neural networks, e.g., spectral, and spatial have been introduced to address various real-world problems such as B2B \cite{ZhanB2019, Bruna2014}. A work in \cite{Henaff2015} proposed a GCN model called ChebNet based on Chebyshev polynomial \cite{Defferrard2017} which later is improved by Kipf et al. \cite{Kipf2017} using first-order spectral propagation. Spatial graph convolution networks demonstrate better generalization ability by aggregating the node neighborhood information. Due to learnable parameters independent of graph size, these spatial networks generalize well across different networks \cite{Duvenaud2015}. In another work in \cite{Zhuangc2018}, the diffusion convolutional network model (DCCN) computes the node receptive field based on a node's diffusion transition probability. A work in \cite{Sajjad2019} adopts the random walk for the convolutional operation of neighbor node feature aggregation. The performance efficiency of this model is limited to smaller graph networks due to a preset random walk length.

GraphSAGE is a spatial GCN that enables representation learning \cite{Hamilton2017}. This simple GCN model is scalable and performs exceptionally well with dense and dynamic graphs by aggregating neighborhood node features. It can use either mean, long short term memory (LSTM), or pooling function to perform the node aggregation. The aggregated node features are propagated through a neural network for prediction or classification. MoNET generalizes the graph learning techniques by integrating spatial and spectral graph convolution approaches \cite{Monti2016}. It uses a parametric kernel on pseudo coordinates that corresponds to a set of neighbors of a particular node to learn sharable features. The performance efficiency of MoNET is promising as compared to other graph-based deep learning models.

The shortcomings of GCN models are handled by using a self-attention mechanism \cite{Velick2018}. This enhancement to GCN is called graph attention network (GAN). The attention technique identifies important nodes in a variable size input and performs learning based on convolution operation. The hidden features of each node are extracted using a self-attention mechanism. The proposed model learns well on unseen complex graphs.  Another version of GAN is developed using a multi-head attention mechanism \cite{Adams2018, Chaudhari2019}. The model can perform parallelized self-attention by assigning different attention priorities on a different set of nodes. Despite its stability, one of the major drawbacks of the self-attention mechanism is its limitation to small datasets. 

The graph convolution neural networks are used for customers' default prediction based on the customer-supplier graph network \cite{Martnz2019}. The proposed work developed a risk assessment model on real-world customer/supplier networks using topological graph metrics like clustering co-efficient, node degree, and page rank coupled with GCN. The customer-supplier graph model is constructed by defining node relationships based on contact sharing, financial flow, and other relevant features in the domain.  The default label is predicted on 168305 company nodes with 310084 edges. The GCN model achieves better performance in contrast to logistic regression. A work in  \cite{Matsunaga2019} proposed stock market prediction using graph neural networks based on the company knowledge graph.  The generalization ability of the GNN model is validated using rolling window-analysis-based-backtesting. It adopts well on a temporal GCN model to deal with time-series data. An additional LSTM layer is stacked on top of the GCN model to predict the stock market price based on temporal features from historic data.

In recent work, \cite{Kim2019}, a graph attention network is proposed for stock-flow prediction. It uses a hierarchical GAN model to aggregate node features based on relationships. The diffusion mechanism embedded in the model selectively extracts node features based on relationship weights generated through high-level graph representation.  Stock prediction and market index classifications are performed using a task-specific layer defined on top of the GAN model.  Additional features from the historical data are extracted using LSTM combined with gated recurrent units (GRU). The stock prediction task is performed using a linear transformation layer, whereas the market index is predicted using graph pooling.  The proposed model outperforms the GCN model and stresses the importance of relationship types in a graph database model. A GCN-based supply-demand prediction is discussed in \cite{Kim2019}. It predicts the hourly demand for bikes in a public-bike sharing business environment based on temporal and spatial properties. The predictive power of the proposed model is responsive to sudden changes in global features like weather conditions.

A detailed investigation of the above work for graph database and graph-based deep learning in the B2B domain reveals that only a few studies have considered graph-based deep learning based on the graph database. To the best of our knowledge, our work is the first effort to implement graph convolutional networks for CRM analytics.

\section{Description of Dataset} \label{dataset}
This work considers the dataset from the B2B dataset, a real-world dataset, from the Salvirt Limited \cite{b2b2020}. The selected dataset does not contain any sensitive information regarding clients, business products, and strategies.  The dataset consists of anonymized information of sale instances from a genuine organization trading in software solutions and services globally.  B2B marketing and sales process follows an auxiliary approach and structural procedure for setting up a connection among the clients and plays a significant role in CRM. To address the gap in the accessibility of informational indexes identified with B2B deals, the selected dataset can be used for the logical analysis of the CRM domain using data analysis and machine learning.

The dataset contains 448 instances or rows with 23 attributes with the labeled sales status column. Initially, the raw dataset does not include a unique feature. An additional column named 'sales\_enquiry\_id' is created as a part of data preprocessing. The newly created column corresponds to a unique sales id for each sale attempt. The unique sale ID is meaningful to design the graph database model that abstracts as a selected dataset in CSV format. After data preprocessing, the selected dataset is free from missing values and noisy data. All features in the dataset have the object data type that represents categorical variables. After preprocessing, the dataset contains 449 rows with 24 attributes or columns that include one labeled sales status column and the newly created $'sales\_enquiry\_id'$ column with 448 unique values. A detailed description of each dataset attributes is presented in Table \ref{tab1}.

\begin{table*}[t]
		\renewcommand{\arraystretch}{1}%
		\centering
		\caption{B2B sales dataset feature description}
		\label{tab1}
		\begin{tabular}{ |p{2cm}|p{3cm}|p{2.5cm}|p{5.5cm}| } 
			\hline
			\rowcolor{lightgray}
			No.
			& Feature Name & Feature Code
			& Feature Description
			\\\hline
			1
			& Product name
			&  Product
			&  Offered product code.
			\\\hline
			2 
			& Seller name
			& Seller
			& Name of the in-charge seller
			\\\hline
			3
			& Authority
			& Authority
			& Authority level at client side
			\\\hline
			4
			& Company size
			& $Comp\_size$
			& Size of the client or company
			\\\hline
			5
			& Competitors	
			& Competitors	
			& Competitors for a sale
			\\\hline
			6
			& Purchasing department	
			& $Purch\_dept$	
			& purchasing department involved
			\\\hline
			7
			& Partnership	
			& Partnership	
			& Product is being sold in partnership
			\\\hline
			8
			& Budget allocation	
			& $Budgt\_alloc$	
			& Reservation of budget by the client 
			\\\hline
			9
			& Formal tender	
			& $Forml\_tend$	
			& Tendering procedure 
			\\\hline
			10
			& RFI	
			& RFI	
			& Request for Information
			\\\hline
			11
			& ROF	
			& RFP	
			& Request for proposal
			\\\hline
			12
			& Growth of a client	
			& Growth	
			& Growth status of the client
			\\\hline
			13
			& Positive statements	
			& $Posit\_statm$	
			& Client 's positive attitude
			\\\hline
			14
			& Source	
			& Source	
			& Source of sales inquiry
			\\\hline
			15
			& Client	
			& Client	
			& Type of a client
			\\\hline
			16
			& Scope  clarity
			& Scope			
			& Clarity of implementation scope
			\\\hline
			17
			& Strategic deal	
			& $Strat\_deal$		
			&  Deal with a strategic value
			\\\hline
			18
			& Cross sale		
			& $Cross\_sale$		
			& Different product sold to a client 
			\\\hline
			19
			& Up scale	
			& $Up\_sale$		
			& Upgrading or increasing existing product	
			\\\hline
			20
			& Deal type	
			& $Deal\_type$	T	
			& ype of a sale or business requirement
			\\\hline
			21
			& Needs defined
			& 	$Needs\_def$			
			& Clearly expressed needs of a client
			\\\hline
			22
			& Attention to client	
			& $Att\_t\_client$		
			& Attention/importance to a client	
			\\\hline
			23
			& Status	
			& Status		
			& An outcome of sales opportunity	
			\\\hline
			24
			& Sales enquiry ID 		
			& $sales\_enquiry\_id$		
			& Unique sales inquiry ID
			\\\hline
		\end{tabular}
		
	\end{table*}

\section{CRM Graph Datbase Design} \label{crmmodel}

\begin{figure*}[ht]
\center
    \includegraphics[height=70mm]{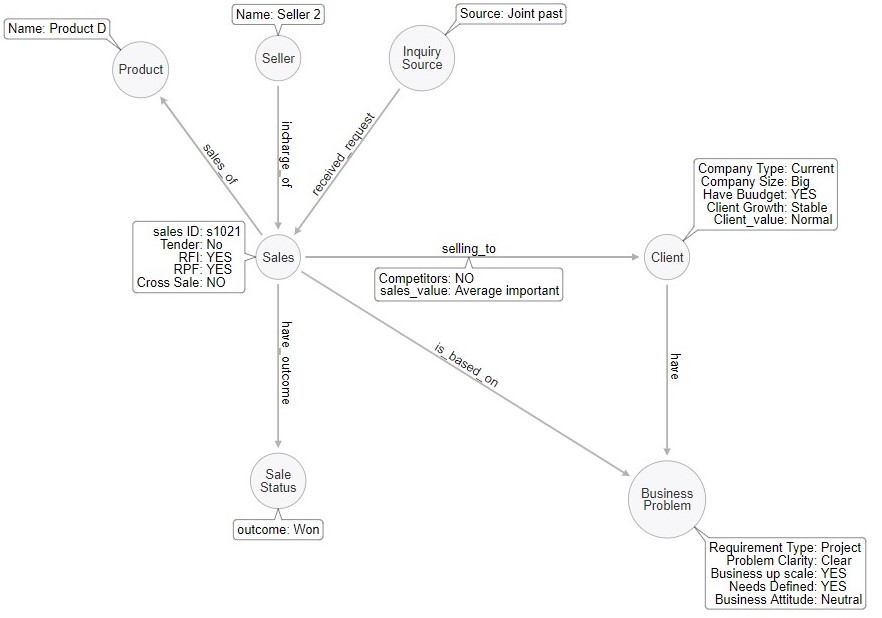}  
    \caption{EDA-Graph database design.} \label{EDA} 
\end{figure*}

To realize the full benefits of graph-based deep learning, in this section, we present the design of two graph databases for the CRM called EDA-Graph database and GCN-Graph database, respectively. These database designs capture all the meaningful information from the CRM domain dataset \cite{b2b2020}. The  EDA-Graph database is useful for query-based data mining and exploratory analysis to identify pattern recognition and interactive query answering abilities. The second graph database model called GCN-Graph database is extracted from the interconnectivity of ‘sale\_enquiry\_id’ nodes. Both the selected graph database models are the outcome of testing various graph database designs evaluated based on data mining efficiency and querying capability. In both, the graph databases,  graphs, nodes, and relationships are primarily modeled entities according to the selected use case. The EDA-Graph model is defined using 7 labels and 7 different relationships between node entities. The EDA-Graph database model is presented in Figure \ref{EDA}.  Based on the use case, nodes represent dataset entities or objects. In the graph, the start node and target node depicts the direction of the relationship.  Neo4j has equal traversal performance in both directions that can query the association without specifying any direction \cite{robinson2020}.

\begin{figure}[ht] 
  \center
    \includegraphics[width=.3\textwidth]{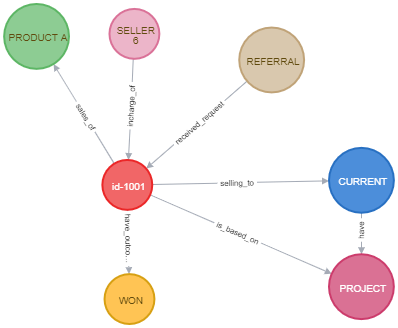}   
    \caption{EDA-Graph model example: sale id: id-1001.} \label{EDA1001} 
\end{figure}

This work selects 7 features from the selected dataset to represent real-life entities as nodes using appropriate labels.  The remaining features are assigned as node attributes of each label or as relationship property. For the sales label,  the sales\_enquiry\_id column is assigned as the first node attribute that creates 448 nodes of unique sale inquiries. Similarly, for the label 'product',  14 unique product labels are created by assigning the product column in the dataset as the first node attribute. Other features, i.e., 'Forml\_tend', 'RFI', 'RPF', and 'Cross\_sale' that define sales constraints are defined as node attributes of that label. Each sales node contains its corresponding dataset values described by the assigned node attributes. Similarly, the client label is described by assigning the 'client' column as a node value. Other dataset features like 'Comp\_size', 'Budget\_alloc, and 'Growth' that describe the individual property of a client are set as node attributes. The ‘Competitors’ and ‘Strat\_deal’ columns of the dataset are assigned as the attributes of the relationship ‘selling\_to’ is defined between the sale and client node. At the implementation level, constraints, nodes, relationships, and properties are defined using Neo4j cypher query. The dataset is imported and integrated with the graph database model using Neo4j user interface. Figure \ref{EDA} illustrates the EDA-Graph database design. Figure \ref{EDA1001} illustrates the graph model of sales id-1001 extracted using the Cypher query language. Each label and related nodes are indicated using different colors. The number of nodes under a specific label is also evident in Figure \ref{EDA1001}.

\begin{figure*}[ht] 
   \center
    \includegraphics[height=70mm]{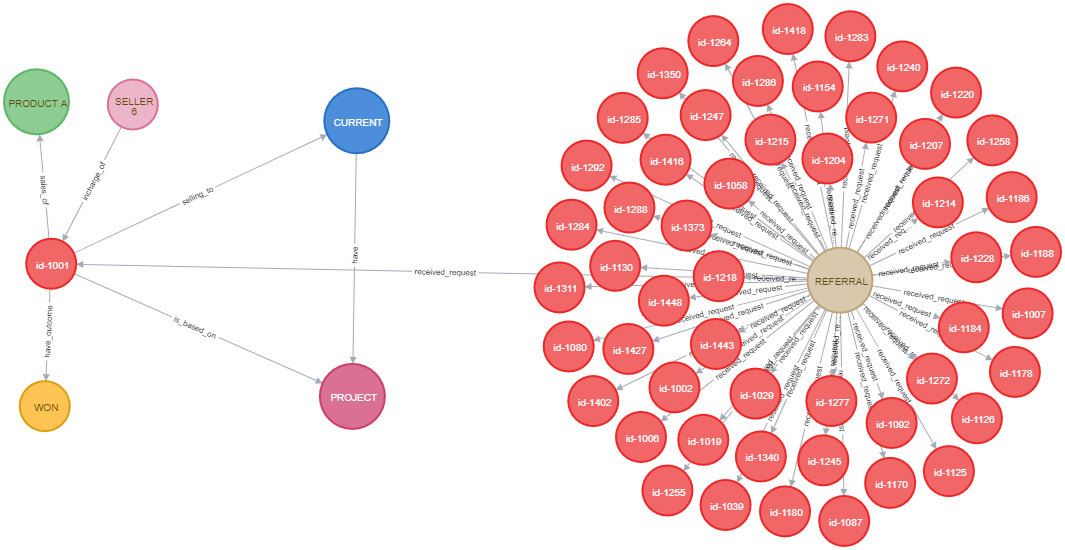}  
    \caption{EDA-Graph model- sale ID: 1001 and all sale ID's of enquiry = REFERAL.} \label{rfereeda} 
\end{figure*}

The value corresponding to each label for a node is given inside the node circle.  As shown in Figure \ref{EDA1001}, the sales id-1001 denotes the sale attempt of product A to an existing client based on business requirements. The node ‘seller 6’ represents a seller who is in charge of the sales, and the source of the sale inquiry is the node 'referral'. From the Figure, it can be observed that sales id-1001 has a successful sales outcome and satisfy project requirements. Figure \ref{rfereeda} presents the graph model of the sales id-1001  along with a list of all other sales nodes whose inquiry source is also 'Referral'. The EDA- Graph model can extract direct visualization instances using the Cypher query.

\begin{figure}[ht] 
\centering
    \includegraphics[width=.07\textwidth]{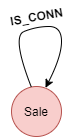}   
    \caption{Single label graph schema.} \label{gcnnode} 
\end{figure}
\begin{figure*}[ht] 
 \center
    \includegraphics[height=70mm]{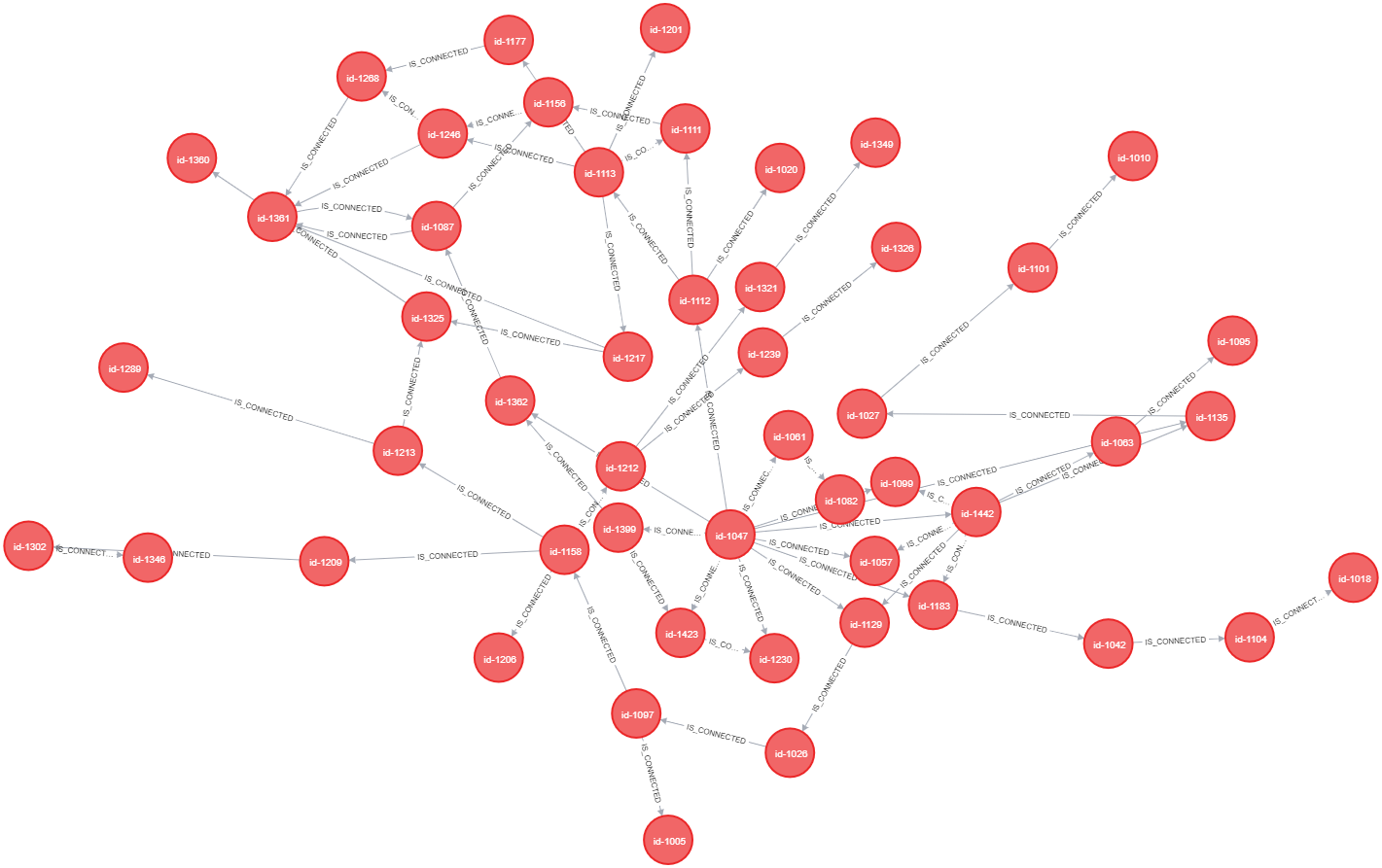}  
    \caption{GCN-Graph model: Sample of 50 sales nodes.} \label{gcnmodel} 
\end{figure*}

The second graph model - the GCN-Graph database is extracted from the first graph model by defining the interconnectivity of ‘sale\_enquiry\_id’ nodes. Figure \ref{gcnnode} depicts the single label graph schema visualization of the GCN-Graph database. The arrow indicates the connection nodes of the sales label. A densely connected graph network of sales nodes is created using 'Up\_sale', 'Client', 'Competitors', 'Product', and 'Seller' attributes of the dataset. These features are selected based on their importance and influence on the sales status prediction, and we use these features to define the relational connectivity between the nodes. The newly created GCN-Graph database consists of 367 nodes and 423 relationships. Figure \ref{gcnmodel} depicts a 50 node instance of the GCN-Graph database. We have used this model to extract the node list and the attribute list to implement the GCN model.

\section{Feature Engineering} \label{fe}

We have transformed the data into an appropriate format, i.e., data type 'status' to an integer. Similarly, the class labels 'Won' and 'Loss' is changed to 1 and 0 binary values. We have also dropped the 'sale\_enquiry\_id’ column that does not influence the label classification and prediction.  The dataset is partitioned into train and test sets with an 80:20 ratio after one-hot encoding on all categorical features. The machine learning and deep learning models are trained using X\_train, y\_train datasets, and the performance is validated against the test dataset. For a steady and unbiased execution, the models are tested on unseen data to establish their generalization ability. The evaluation results demonstrate the model classification with generalization ability. In our work, training and testing steps are performed on independent subsets of data.

Datasets required for the GCN model implementation are extracted from the GCN-Graph database model exported in CSV file format from the Neo4j interface. The work has created two new datasets called 'CRM.edgelist' and CRM.attributes called as the adjacency matrix and the label matrix. The dataset 'CRM.edgelist' defines the node to node network connectivity or relationship in the GCN-Graph database. The 'CRM.attributes' file in CSV format contains the list of sales nodes, node label for performance evaluation, and a pair of labeled nodes (one node with label 'Won' and another node with label 'Loss') for GCN model training. The feature matrix of each node is extracted dynamically.

We have selected two separate nodes with sale status as 'Won' and 'Loss', and flag them separately in the dataset for training purposes. We have used important dataset features to define the GCN-Graph model, hence, we have excluded the node feature matrix to avoid bias in the prediction. Instead of using a feature matrix corresponding to dataset sales instances, we have extracted and used a set of new graph specific characteristics from the network and used it as the feature matrix in the model. The graph specific features are extracted using both Neo4j graph data science extension and python NetworkX library. The graph features used in the GCN model are as follows:

\begin{itemize}
\item   \textbf{Page Rank \cite{Brin1998}} algorithm is used to measure the importance of each node in a graph based on the number and influence of each source node. Higher the page rank of a node in the network, the stronger its impact on the network.

\item \textbf{Closeness Centrality \cite{Sabidussi1966}} of a node returns the reciprocal of the shortest path distance from a node to all other nodes in the network. It defines the information spreading efficiency of a node in the graph. The higher the closeness centrality, the node is more near to all other nodes in the network.

\item \textbf{Identity Matrix \cite{Pipes1963}} provides node features directly, i.e., there are no dataset instances based features defined for a node. We define an identity matrix of the same size as the adjacency matrix t as a one-hot encoded categorical variable representation of each node.

\item \textbf{Clustering Coefficient \cite{Watts1998}} of a node in a network is the measure of the likelihood of interconnection between neighbors of that node. We measure the local clustering coefficient of a node to measure the embedded quality of each node.

\item \textbf{Shortest Path } of a node in a network is the shortest path from that node to the specified target node. In this use case, we are eliciting the shortest path of each node in the network by assigning the labeled training nodes as targets. The extracted feature matrix is an $N \times 2$ matrix containing the shortest path of each node to two separate target nodes, i.e., nodes with labels 'Lost' and 'Won'.

\item \textbf{ Eigenvector Centrality \cite{Newman2008} } of a node is the influence of that node in the graph. The relationship of a node with a high scoring node is a reason to get a more eigenvector score than its connection to a low scoring node.

\end{itemize}

\section{GCN-based Analytics} \label{gcnn}
In this work, we have applied graph-based analytics using GCN on the graph database model from Section \ref{crmmodel} using Neo4j. We have compared the performance of the GCN model based on the graph database model with CNN, ANN, and tree-based RF techniques.

GCN \cite{Bruna2014} is an advanced neural network model introduced for graph analysis that can perform convolutions on the graph model and aggregates the node information from the neighborhood. The expressive power of GCN in graph representation makes it a powerful tool for graph analysis.  The goal of GCN is to learn a function of signals or features on a graph G = (V, E) that takes a feature matrix and an adjacency node connection matrix as input. The GCN-based analytics for the graph database model extracted in Section \ref{crmmodel} is given in Algorithm \ref{alg1} and Algorithm \ref{alg2}.

Given a graph network G(V, E) where V represents nodes connected with edges E.  Algorithm \ref{alg1} extracts feature matrix $F$ corresponding to the graph G with the $F \times N$ dimension where $N$ denotes the number of nodes and $F$ is the number of features. Line 1 in Algorithm \ref{alg1} generates an adjacency matrix A of size $N$ denoted as $ \hat{A}$. The adjacency matrix contains information on the weighted edges. $A_{ij}$ is set to 1 if an edge exists between $i$ and $j$ and otherwise $0$. Line 4 to 5 in Algorithm \ref{alg1} generates an $N \times L$ binary matrix $Y$ where $L$ represents the number of classes. The generated matrix is a single instance of each label. The labelled matrix corresponding to the test set is used later for performance evaluation.

The adjacency matrix A and feature matrix X are fed to the forward propagation equation. The adjacency matrix is passed to the forward propagation in Equation\ref{eq1} after adding self-loop to include the features of the present node along with neighboring node features. The forward propagation in  Equation\ref{eq1} generates each node's feature as a summation of features of the neighboring node along with its features. The simple forward propagation equation defines the addition of self-loop with the adjacency matrix using the identity matrix and is given as follows in Equation \ref{eq1}:

\begin{equation} \label{eq1}
\hat{A} = A+1
\end{equation}
             
Line 5 to 7 in Algorithm \ref{alg1} normalize the features using the inverse degree matrix (Chung et al., 2003) where $D^{-1}$ is the inverse degree matrix and is expressed as given in Equation \ref{eq2}.

\begin{equation} \label{eq2}
f(X,A) = D^{-1}\hat{A} X 
\end{equation}

Each hidden layer in GCN is represented as the propagation rule as given in Equation \ref{eq3}. In the Equation \ref{eq3}, $H^{i}$ represents the $i^{th}$ hidden layer and $f$ is the propagation function where $H^{0}=X$. The $H^{i}$ is the propagated $i^{th}$ row in a feature matrix $N \times F^{i}$ and is the based on the spectral rule given Equation \ref{eq3}.

\begin{equation} \label{eq3}
H^{i} = f(H^{i-1}, A)
\end{equation}

The propagation function is given in Equation \ref{eq4}:
\begin{equation} \label{eq4}
f(H^{i-1}, A) = \sigma (D^{-1}(\hat{A})H_{i}W_{i})
\end{equation}

Here $W^{i}$ is the weight matrix for layer $i$ and $\sigma$ represents the activation function ReLu. Based on Kipf and Welling's spectral propagation rule  \cite{Kipf2017}, we can rewrite this equation as Equation \ref{eq5} and illustrated in Line 6 to 7.

\begin{equation} \label{eq5}
f(H^{i-1}, A) = \sigma (D^{-0.5}(\hat{A})D^{-0.5}H_{i}W_{i})
\end{equation}

\begin{algorithm}
  \SetAlgoLined
 \KwData{Adjacency matrix $A$, Feature matrix $X$, Label matrix $L$}
  \KwResult{Latent feature representation $F_{latent}$}
  $H\Leftarrow$  number of initialized hidden neural network layers 
  
          $ \hat{A} \Leftarrow A+1$ 
           \Comment{Generate adjacency matrix A of size$N \times N$}

        \textbf {function \funccall {GCNmodel}} ( $ \hat{A}$)
         
          $ CON\_Aggregator \Leftarrow \emptyset$
          
           \For {$h=1$ to $H$} {
           For {$I=1$ to $(number\_of\_rows(L))$}
           {$CON\_Aggregator. append (\sigma (D^{-0.5}(\hat{A})D^{-0.5}H_{i}W_{i}))$
            \Comment{Kipf and Welling's spectral propagation rule} 
                  
                  return LogisticRegressor $(CON\_Aggregator)$}
                  
                   }
                   
\caption{GCN with convolution operation} \label{alg1}

\end{algorithm}

\begin{algorithm}
  \SetAlgoLined
 \KwData{crm.edgelist, crm.attributes.csv}
  \KwResult{Sales prediction}
     $network \Leftarrow$  crm.edgelist
    
              $ attributes \Leftarrow$ crm.attributes.csv
           \Comment{Generate adjacency matrix A of size$N \times N$} 
 
       $X\_train, Y\_train \Leftarrow$  (attributes[‘churn’]==’Won’, ‘Lost’)
       \Comment{Setting labeled nodes for training} 
       
            \hspace{4mm}  $X\_test, Y\_test \Leftarrow$  (attributes[‘churn’]==’sales’) 
       \Comment{Setting all unlabeled nodes for testing} 

             $X\_train, X\_test \Leftarrow$  flattern(X\_train, X\_test)
         \Comment{partition date into test and train sets}
         
              $train(GCNmodel, X, X\_train, Y\_train)$
           \Comment{Feed the GCNmodel training data} 
           
          $ predict(GCNmodel, X, X\_test)$ 
         \Comment{Perform predictions} 

             \caption{GCNModel Training and prediction} \label{alg2}

\end{algorithm}

Algorithm \ref{alg2} illustrates the training and testing of the GCN model based on the $F_{latent}$ features generated by Algorithm \ref{alg1}. The Algorithm \ref{alg2} trains the model on the labeled data and propagate the learned features through convolutional layers.
The Algorithm updates the weight matrix using the cross-entropy loss computed on known node labels. The cross-entropy loss backpropagates until an optimal weight matrix is found.  
At each layer, features are aggregated using the non-linear activation function ReLu with layer-wise spectral propagation. Line 7 in Algorithm \ref{alg1} shows that the GCN model is further passed to the logistic regression model to perform binary classification using node features learned using spectral propagation.

\section{Performance Evaluations} \label{pe}
This section evaluates the performance of GCN and compares it with CNN, ANN, and random forest-based ensemble algorithm \cite{Zhu2016}. The used ANN is a feed-forward neural network \cite{LeCun2015}. In our evaluations, we train the ANN model using a one-hot encoding training dataset, and the CNN with 95 training features with 2D convolution operation. 

The performance evaluation uses a wide range of python and graph analysis libraries. The implementation of deep learning models has been carried on NVIDIA GeForce GTX-1050 Ti graphic processing unit. The GPU unit is composed of 766 NVIDIA CUDA cores, 4 GB shared memory (VRAM) and has 1392 MHz clock speed. The implementation has used Python-3.6 and Cypher \cite{Nadime2018} for data analysis, model development, and data mining. Cypher by Neo4j is a declarative graph query language developed for efficient and expressive graph data processing, i.e., graph model creation and graph feature extraction. The exploratory analysis is based on Python and Cypher. GCN model is implemented by using  Keras or MXNet python deep learning packages. Other libraries include network XNetworkX, SKlearn, TensorFlow, py2neo, and APOC.

\subsection{Graph Database Inventory Statistics}
We have presented two two graph database models called an EDA-Graph database and GCN-Graph database in Section \ref{crmmodel}.  The EDA-Graph database abstracts the B2B sales dataset to perform exploratory data analysis and data mining. The GCN model is implemented on top of the GCN-Graph database. The GCN-Graph database represents an interconnected network of the sales nodes corresponding to 'sales\_enquiry\_id' in the dataset. The connectivity between the sales nodes is defined using five dataset attributes, i.e.,   Up\_sale, Client, Competitors, Product, and seller.

The EDA-Graph database design is presented in Table \ref{tab2}. The model consists of 1386 nodes with 7 labels. The total number of relationships between the nodes is 3136 with 7 relationship types. Among the 7 assigned labels, five are unique labels created using Cypher query 'unique constraint' declaration. Unique labels are used to merge the same labeled nodes with similar values during the data import procedure. The size of the database is 2.03MB,  larger than the raw dataset of size 67 KB. The sales label consists of 448 unique sales\_enquiry\_id\_nodes, whereas the product label defines 14 unique products sold by the company. There are 18 sellers represented by the 18 distinct seller nodes in the graph database. The sales inquiry received from the eight different sources is integrated into the database as eight individual nodes with the sale status label with unique values of 'Won' and 'Lost'. These values represent the labelled status column in the CSV dataset.

\begin{table}[t]
		\renewcommand{\arraystretch}{0.5}%
		\centering
		\caption{ EDA-Graph database inventory statistics } \label{tab2}
		
		\begin{tabular}{|p{3.2cm}|p{0.8cm}|p{1.8cm}|p{0.8cm}| } 
			\hline
			\rowcolor{lightgray}
		Parameter name & Value & Unique labels & Count
			\\\hline
			Number of nodes & 1386 & Sale & 448
			\\\hline
			Number of labels & 7 & Product	 & 14
						\\\hline
			
			No. of relationships & 3136 & Seller	 & 18
			\\\hline
			No. of unique labels & 5 &  Source	 & 8
			\\\hline
			
			No. of relationship types & 7 & Sales Status	 & 2 
			\\\hline
					
			Size & 2.03MB  & - & -       
			\\\hline
			
		\end{tabular} 
		
	\end{table}

The inventory statistics of the  GCN-Graph database model are shown in Table \ref{tab3}.  These metadata statistics and the database schema are extracted using Cypher query language. The GCN-Graph database model consists of a single sales label and is composed of 367 sale ID nodes.  There are 423 relationships between sale nodes with one type only. The size of the database is 1.53 MB,  less than the size of the EDA-Graph database. This is due to less number of labels, nodes, and relationships. The average relationship count of the sale label is 2.305 per node. The maximum relationship count is 33 (dense) with a minimum value of 1 (sparse), i.e., a graph with a maximum of 33 nodes and a minimum of 1 node.

\begin{table}[t]
	\renewcommand{\arraystretch}{1}
		\centering
		\caption{ GCN-Graph database inventory statistics }
		\label{tab3}
		\begin{tabular}{|p{3cm}|p{2.5cm}| } 
			\hline
			\rowcolor{lightgray}
	Parameter name & Value 
			\\\hline
			No. of nodes & 367 
			\\\hline
			No. of labels & 1 
						\\\hline
			
			No. of relationships & 423 
			\\\hline
			Min. relationship count & 1.00 
			\\\hline
			Max. relationship count & 33.00
			\\\hline
			Avg. relationship count & 2.305
			\\\hline
			Relationship types & 1 
			\\\hline
							
			Store size & 1.53MB  
			\\\hline
		      Name of Label & Sales
		      \\\hline
				Name of relationship & IS\_CONNECTED
			\\\hline

		\end{tabular}
		
	\end{table}

\subsection{Exploratory Data Analysis}

Exploratory data analysis is performed by using Neo4j Cypher query language. As all the features in the CRM dataset are categorical, hence the quantitative analysis based on statistical measures is irrelevant here. This section presents an analysis of important features with a focus on sales status count. The selected dataset that consists of 448 instances of sales\_enquiry\_id has almost the same number of 'Loss' and 'Won' sale outcome status. Both the EDA-Graph database and GCN-Graph database share the same characteristic. Here the count of sale status classes is evenly distributed. The sales status count distribution of sale ID is listed in Table \ref{tab4} and. Some sales nodes are removed during the GCN-Graph database model creation to avoid the formation of a separate graph network that has no relationship with the main graph network. In the generated graph database, the proportion of sales instances with positive and negative sales outcomes are approximately the same.

The EDA-Graph database consists of 227 successful sale attempts and has 221 sale instances with negative sale outcome. On the other hand,  the GCN-Graph model consists of a total of 367 sale nodes with 177 sale instances with positive sale outcomes, and 190 instances with a negative outcome. 

\begin{table}[t]
		\renewcommand{\arraystretch}{0.5}%
		\centering
		\caption{Frequency table of sales ID based on sales outcome } 
		\label{tab4}
		\begin{tabular}{|p{2cm}|p{2cm}|p{2cm}| } 
			\hline
			\rowcolor{lightgray}
		Sales Status & EDA-Graph Count & GCN-Graph Count
			\\\hline
			Won & 227 & 177
			\\\hline
			Loss & 221 & 190
				\\\hline
		    Total & 448 &  367
					\\\hline	
			
		\end{tabular} 
		
	\end{table}

\begin{figure*}[htp]
  \centering
  \subfigure[Product-Seller-count plot based on sale status.]{\includegraphics[scale=0.4]{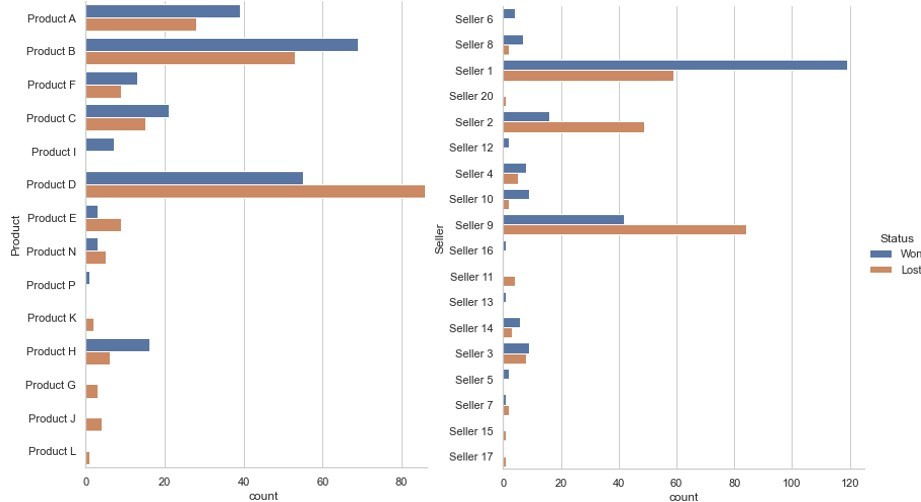}}  \quad
  \subfigure[Competitors, Client, Company size and UpScale versus sale status count.]{\includegraphics[scale=0.4]{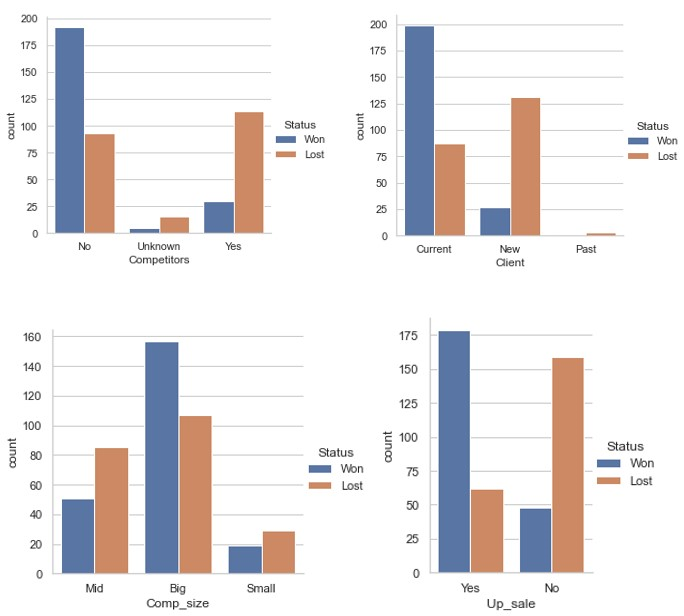}} 
  \caption{Exploratory data analysis} \label{edanlysis}
\end{figure*}

Figure \ref{edanlysis} depicts the count of the seller and product features with the sales status.  It is clear from the figure that Product B and Product D are the two most successful products with a higher sales rate. Product D receives the highest number of sale requests. Despite this, Product B still achieves a higher sales success count. In the Figure, the positive sales status of Product D is less than the negative sales status. Based on the selected dataset and use case, it is evident that the successful sales count of Product B is just below 70 and the number of failed sales attempts is below 60. Products L, J, G, and K have very few inquiries and with no successful sales outcomes. From the seller vs. sales status plot from Figure \ref{edanlysis} (a), we can see that Seller 1 is the most successful salesman in charge. The number of positive sales performed by Seller 1 is just below 120 with less than 60 failed sales outcomes. Seller 2 and Seller 9 are the two other sellers with comparatively better positive sales outcomes than the remaining 15 sellers. The product and seller features are use case specific and do not generalize to other B2B sales attempts. Other than product and seller features, it is also important to consider other external features like innovation, geographic features, and market needs while making this assumption.

Figure \ref{edanlysis} (b) shows features that generalize well in the B2B sale process. These features are important for CRM with a significant effect on sales outcomes. These features are transferable with an ability to generalize in similar business use cases. It can be observed that the success rate of a company with no competitors is higher than others.  A similar trend can be observed in the column 'Client'. For current or existing clients, the sale success rate is twice the count of negative sales outcomes.
In this use case, the sale success count is positively affected by potential clients. For potential clients, the number of positive sales outcomes is higher than the failed sales attempts. From the analysis of the ‘UpScale’ feature from Figure \ref{edanlysis} (b), it is clear that all clients in need to  ‘UpScale’  the products have a positive effect on the sales success count.  To conclude, we can say that the sales attempts with no competitors, existing clients, and clients with upscale requirements have a positive effect or correlation to the sales success rate.

\subsection{GCN Model Evaluation and Analysis }
In this section, we compare the performance of GCN models with RF, CNN, and ANN in terms of precision, sensitivity, specificity F1 score, and Area under the ROC Curve (AUC).  Specifically, we analyze the performance of GCN for each extracted graph features discussed in Section \ref{crmmodel}. Further, we also analyze the performance of GCN on various graph feature combinations. 

\begin{figure*}[htp]
\centering
 \subfigure[ GCN model with baseline models.]{\includegraphics[scale=0.75]{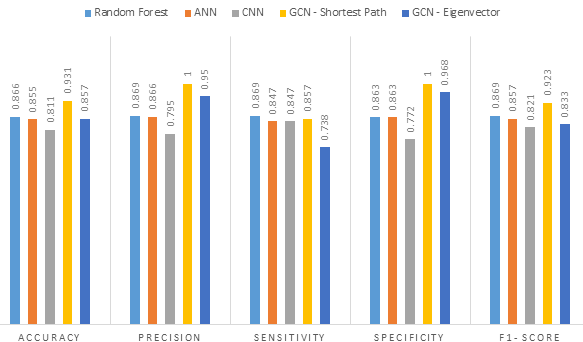}} \label{perfm1}  \quad
  \subfigure[Area under the ROC Curve of GCN, RF, CNN, and ANN models.]{\includegraphics[scale=0.35]{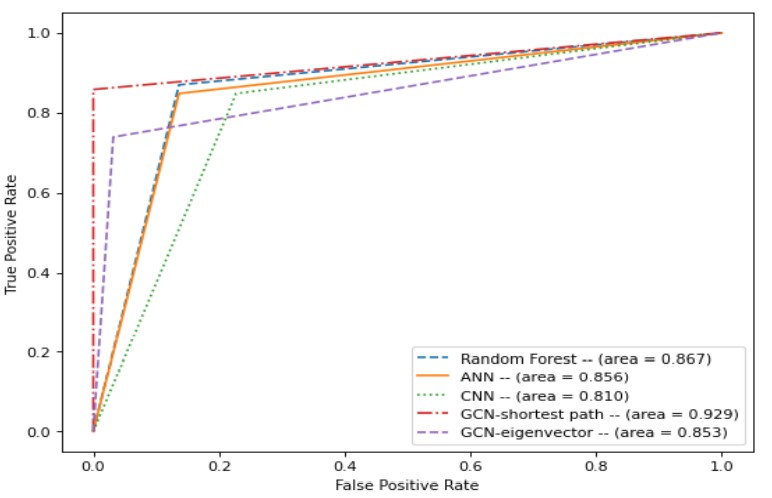}} \label{gcnf} 
  \caption{Evaluation results of GCN model} \label{gcnanalysis}
  
\end{figure*}

In the confusion matrix, the negative label represents the failed sales outcome as '0' and the positive class as '1' for successful sales. Precision is the measure of the number of correctly classified sale statuses as positive out of all positive predictions.  Sensitivity also called recall is the percentage of actual positive sales outcomes that are predicted correctly. Specificity represents the ratio between total correctly classified high priority negative sale instances to total negative sale outcome prediction.  F1-Score is the harmonic mean of both precision and recall and represents a model's classification ability. In evaluations, we plot the ROC curve by using sensitivity on the x-axis and 1-specificity on the y-axis.
Figure \ref{gcnanalysis} (a) shows the performance comparison of GCN-Shortest Path and GCN-Eigenvector with RF, CNN, and ANN. The GCN-Shortest Path performs better than all other models for accuracy, precision, specificity, and F1-Score. The sensitivity of the GCN models is lower than the RF model with a value of 0.869.  The GCN-Shortest Path model demonstrates 100\% precision and specificity with a sensitivity value of 0.857.  It is evident from the results that both the versions of the GCN models outperform RF, ANN, and CNN for precision and sensitivity. The GCN-Shortest Path model outperforms all models with an accuracy of 0.93. On the other hand, the accuracy observed by the RF and GCN-Eigenvector model is 0.86, 0.85, respectively. ANN outperforms GCN-Eigenvector with an F1 score of 0.85. It means ANN generalizes better compared to the GCN-Eigenvector model. Results in Figure \ref{gcnanalysis} (a)  reveal that the sensitivity of both the CNN and ANN is 0.847 and is better than the GCN-Eigenvector model. It is observed from the Figure that the RF outperforms other models with an F1-Score of 0.869.

The number of correct and incorrect sales outcomes prediction of the models is summarized in Table \ref{tab5}. The true positive rate of all models except GCN-Eigenvector is above 80\%. The two GCN models have the best true negative value and are successful at identifying the sale attempts with negative results correctly. The best model, GCN-Shortest Path has a true negative rate of 1.00 and a true positive rate of 0.86. The true positive rate of the GCN-Shortest path is slightly less than the true positive rate of the RF.  The RF model has the highest success rate for predicting true positive sales outcomes. The performance of both the GCN models is better than all other models to successfully classify the sales outcomes.

\begin{table}[t]
	
		\centering
		\caption{ Confusion matrix of GCN, RF, CNN, and ANN models }
		\label{tab5}
		\begin{tabular}{|p{2.5cm}|p{1cm}|p{1cm}|p{1cm}|p{1cm}| } 
			\hline
			\rowcolor{lightgray}
Model & True Negative & False Positive & False Negative & True Positive
			\\\hline
			Random Forest	& 0.86	& 0.14  &	0.13 &	0.87 
			\\\hline
	ANN &	0.86 &	0.14 &	0.15 &	0.85
	\\\hline
	CNN &	0.77 &	0.23 &	0.15 &	0.85
						\\\hline
						
	GCN-shortest path &	1 &	0 &	0.14 &	0.86					
							\\\hline
							
							GCN-Eigenvector	& 0.97 &	0.03 &	0.26 &	0.74
						\\\hline

		\end{tabular}
		
	\end{table}



Figure \ref{gcnanalysis} (b) shows the ROC curve of all the models. The Figure shows that the GCN-Shortest Path achieves the highest AUC value of 0.929.  It can be observed from the Figure that the GCN-Eigenvector underperforms RF and ANN  with an AUC value of 0.853. These results are obtained from a balanced dataset for training and testing.

In our selected use case relevant to the CRM domain, prediction of both true positive and true negative is equally important to devise sales and marketing strategies.  Here, the GCN-Shortest Path model demonstrates the best performance to predict all true and negative sales outcomes successfully. RF is the best model to predict true positive values.  For an appropriate business model selection, we can consider both the F1-Score and accuracy of models. Based on GCN model evaluations and comparison with other models, it is clear that the GCN-Shortest Path is the best fit model for B2B sales outcome prediction. While in terms of overall stability, the RF model is also a desirable option for deployment.       

The better performance of the GCN model is attributed to the node to node relationship among sales nodes. Node-wise relationship between sales nodes is a strong influencing parameter in terms of its community or cluster behavior. The results reveal that the GCN model that learns by performing spectral convolutions on a set of nodes can extract meaningful features and attributes from neighboring nodes. Here, extracted features like shortest-path and eigenvector centrality values are strong influencing attributes of each node. In terms of convolutional learning, the GCN model can extract insights from the neighboring nodes. Using GCN, it is possible to process non-obvious features like shortest path, eigenvector centrality, and other hidden graph features. These graph features show significant impact on the classification performance  compared to the dataset features using traditional dataset or relational database.

\subsection{Analysis of GCN Model with Graph Features}
\begin{figure*}[htp]
  \centering
  \subfigure[Performance evaluation of all GCN experimental model.]{\includegraphics[scale=0.6]{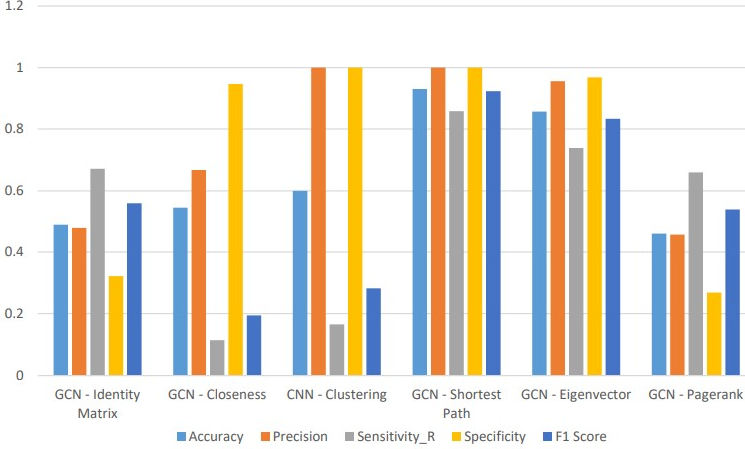}}  \quad
  \subfigure[GCN based on feature count.]{\includegraphics[scale=0.85]{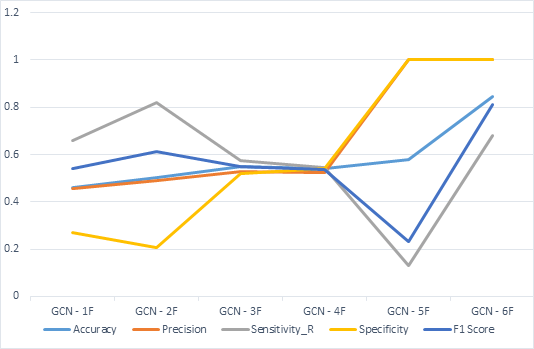}}  
  \caption{Evaluation results of GCN model} \label{fgcnanalysis}
   
\end{figure*}

In this section, we analyze the performance of GCN with various graph features and their combinations, i.e., GCN-Shortest Path, GCN-Eigenvector centrality, GCN-closeness centrality, GCN-Identity Matrix, GCN-Clustering Node value, and GCN-Pagerank. These models perform better in terms of accuracy and F1-score.  Figure \ref{fgcnanalysis} (a) shows the performance of GCN for various performance metrics. It can be observed from the Figure that the precision and specificity of the GCN-clustering model is 100\%, however, with a  compromise on sensitivity. 

GCN-clustering model successfully identifies all true negative sales outcomes, however, the model fails to classify true positive values. GCN-shortest path is the only model with a good true positive rate. Among all the models, the GCN-Closeness and GCN-Clustering models have the least F1-Score due to poor ability to predict the true positive sales outcomes. It is worth mentioning that the performance efficiency GCN-Shortest Path model experiences fluctuations. During experiments, on some rare occasions, we observed instability in the performance of the GCN-Shortest Path model. It is due to the significant difference in the smallest and highest shortest path count of nodes to the target and is related to spectral convolutional random weight initialization of neurons.

The better performance of the GCN-Shortest Path model is due to the shortest path feature extraction procedure. We use labeled nodes as a target to calculate the shortest path distance of each node. It means that the shortest path graph feature has a stronger impact on node classification.  Eigenvector centrality of a node allows the high-scoring nodes to contribute more to the score of the node in question compared to the equal connections to low-scoring nodes.  We also compare the performance of GCN models using different combinations of graph features as listed below:

\begin{itemize}
\item GCN-1F:PageRank
\item GCN-2F:PageRank,Identity Matrix
\item GCN-3F:PageRank,Identity Matrix, Closeness
\item GCN-4F:PageRank, Identity Matrix, Closeness, Cluster.
\item GCN-5F:PageRank, Identity Matrix, Closeness, Cluster, Eigenvector
\item GCN-6F:PageRank, Identity Matrix, Closeness, Cluster, Eigenvector, Shortest Path

\end{itemize}

Figure \ref{fgcnanalysis} (b) shows the performance results of GCN with various graph feature combinations. The accuracy and F1-score of the GCN-5F model are only 0.58 and 0.23, respectively which is lower than the GCN-Eigenvector model. The accuracy and specificity of the model increase with an increase in the number of features. We can also witness a sudden rise in all performance parameters after adding eigenvector and shortest path features to the model especially after adding the shortest path. From the experiments, we can identify that features like PageRank, Closeness Centrality, Cluster node value, and Identity Matrix do not significantly affect the graph network model.

A close analysis of the GCN model with various feature combinations shows that the performance of the GCN model does not solely depend on the number of features. From the analysis, it is also clear that the performance of the GCN model with the Shortest Path and Eigenvector is less than the GCN model without these features. It means that the addition of features negatively impacts the performance efficiency of GCN models.

\section{Conclusion and Future Works} \label{conc}
This work focuses on the graph database design and graph-based deep learning to analyze B2B CRM dataset. We designed two distinct graph database models based on B2B sales dataset using the Neo4j.  On top of this graph database, we applied EDA approaches to extract meaningful insights. Specifically, we used two variants of GCNs to predict sales on graph databases. The performance of GCN is evaluated and compared with RF, CNN, and ANN-Feed forward network. Our results and comparison reveal that the  GCN-Shortest Path model outperforms all other models in terms of accuracy, F1-Score, precision, and sensitivity.  This suggests the  GCN-Shortest Path model as the best model for classification and generalization. However, its sensitivity with a value of 0.857 is lower than the RF with a value of 0.869. The GCN-Shortest Path and GCN-Eigenvector models observe the highest precision and specificity, which implies that the GCN models are successful at identifying negative sales outcomes correctly.  

Performance evaluation based on ROC plots also validated the outstanding performance of GCN models with a value of 0.924. Here GCN Shortest Path model has the best performance to predict all true negative sales successfully, however,  the RF model outperforms the GCN model in terms of true positive prediction success rate. Further, experiments on the GCN model based on various graph features show that the performance of the GCN-Shortest Path and Eigenvector model is better than the GCN model based on other graph features.  This reveals the effect of feature correlations on the performance of the GCN model.

The results show that the performance of GCN models is not affected by the varying number of features. The overall performance of the model is decreased even after adding the Shortest Path and Eigenvector feature values into the models. It indicates the negative influence of additional extracted features. Overall results indicate that a simple integration of the B2B CRM domain into a graph database improves graph-based data mining and graph analytics. It will be interesting to incorporate the social network graph model with the GCN to extract valuable insights and to demonstrate the scalability of the Neo4j graph database. Further, it will be valuable to compare spectral GCN model with spatial GCN models \cite{Hamilton2017}.


%

\ifCLASSOPTIONcaptionsoff
  \newpage
\fi



%

%
\begin{IEEEbiography}[{\includegraphics[width=1in,height=1.25in,clip,keepaspectratio]{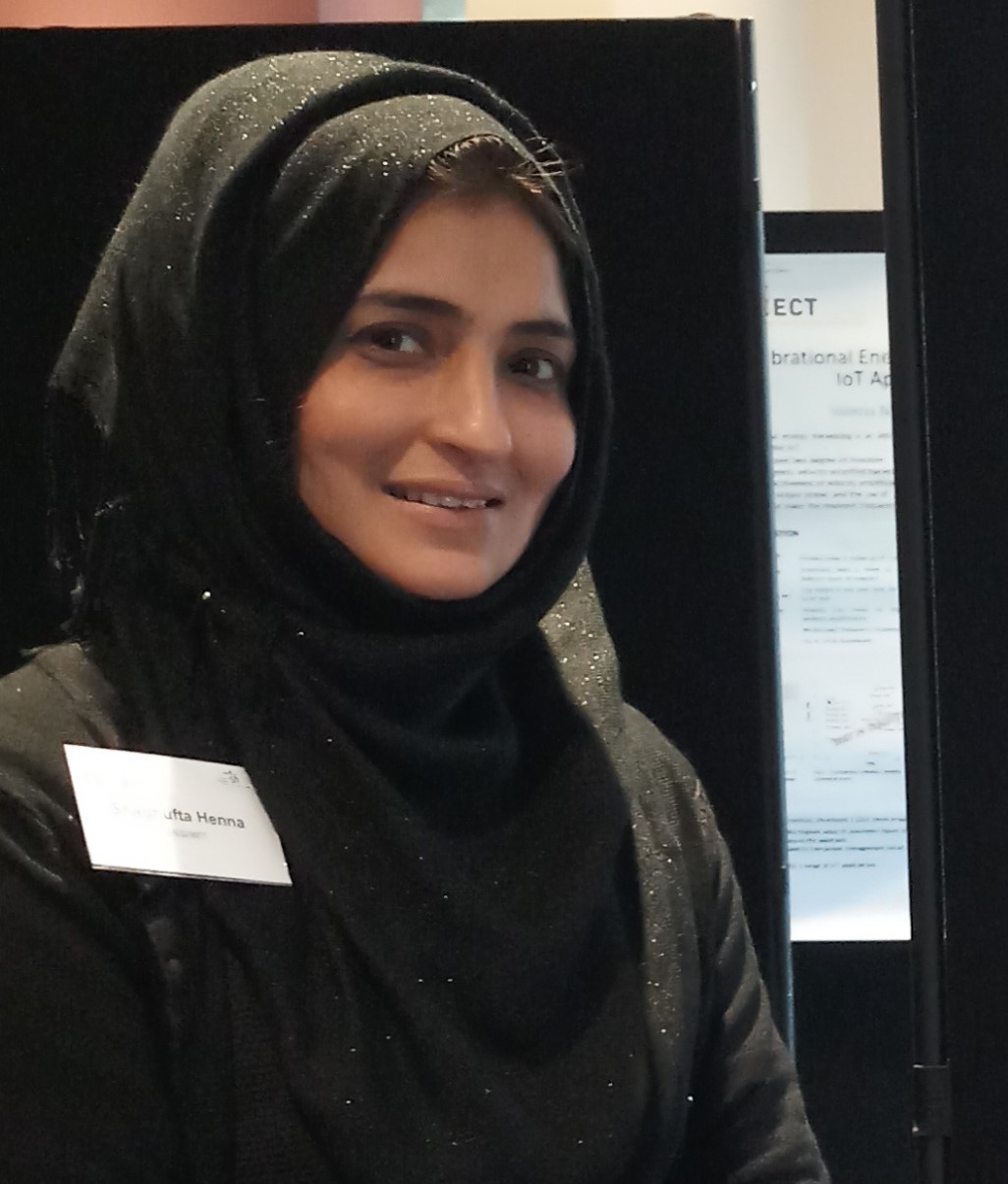}}]{Shagufta Henna} is an assistant lecturer with the Letterkenny Institute of Technology Co. Donegal, Ireland. She was a post-doctoral researcher with the Waterford institute of technology, Ireland. She received her doctoral degree in Computer Science from the University of Leicester, UK in 2013. She is an Associate Editor for IEEE Access,EURASIP Journal on Wireless Communications and Networking, IEEE Future Directions, and Human-centric Computing \&bInformation Sciences, Springer. Her current research
interests include big data analytics, deep learning, and machine learning-driven network optimization.
\end{IEEEbiography}

\begin{IEEEbiography}[{\includegraphics[width=1in,height=1.25in,clip,keepaspectratio]{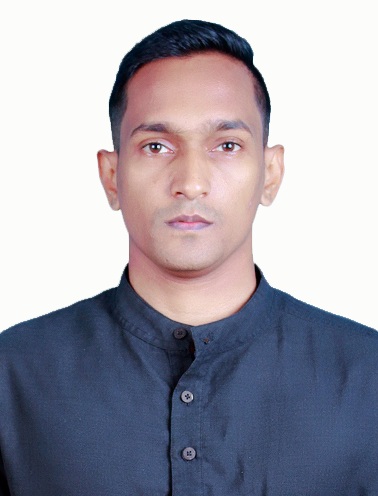}}]{Shyam Krishnan Kalliadan} is Masters student of Big Data Analytics with the Letterkenny Institute of Technology Co. Donegal, Ireland. His current research interests include big data analytics and machine learning.
\end{IEEEbiography}




\end{document}